# REAL-TIME RECOGNITION OF YOGA POSES USING COMPUTER VISION FOR SMART HEALTH CARE

**Abhishek Sharma**
ECE Dept.
The LNMIIT, India

**Yash Shah**
ECE Dept.
The LNMIIT, India

**Yash Agrawal**
ECE Dept.
The LNMIIT, India

**Prateek Jain**
SENSE
VIT AP University, India

January 19, 2022

## ABSTRACT

Nowadays, yoga has become a part of life for many people. Exercises and sports technological assistance is implemented in yoga pose identification. In this work, a self-assistance based yoga posture identification technique is developed, which helps users to perform Yoga with the correction feature in Real-time. The work also presents Yoga-hand mudra (hand gestures) identification. The YOGI dataset has been developed which include 10 Yoga postures with around 400-900 images of each pose and also contain 5 mudras for identification of mudras postures. It contains around 500 images of each mudra. The feature has been extracted by making a skeleton on the body for yoga poses and hand for mudra poses. Two different algorithms have been used for creating a skeleton one for yoga poses and the second for hand mudras. Angles of the joints have been extracted as a features for different machine learning and deep learning models. among all the models XGBoost with RandomSearch CV is most accurate and gives 99.2% accuracy. The complete design framework is described in the present paper.

## 1 Introduction

Yoga, whenever this word comes into our mind, it is imagined as an exercise but, this kind of exercise should be performed by everyone with proper knowledge of the significant posture. We need an appropriate instructor for it but, this is not so easy to find a good instructor, as well as many of them, do not even afford the instructors. Nowadays, people are working on this to make this thing digitalize so that it is required to pay for a yoga class or hire an instructor to perform yoga. There is a software tool that people can use to perform yoga without any guidance. This software is just like an instructor. Whenever it is required to join any classes or hire an instructor, how the instructor helps. This will be the same as instructor oral guidance to perform any pose. If the postures are wrong, an entity must correct them, and the particular person will perform the correct postures. The basic discussion for yoga therapy is presented in Fig. 1.

## 2 Prior Related Work

In our previous work, the paper suggested the identification of different yoga poses through machine learning techniques and various image processing steps [1]. This paper talks about correcting the yoga pose and detecting yoga mudras. There is some research literature to identify yoga mudras through computer vision techniques, but to the best of the authors' knowledge, detecting yoga mudra and postures altogether is the novel contribution of the work. The work presented here provides a unique feature where the practitioners' posture correction, recognition, and mudra identification can be achieved within a single screen. Different techniques have been developed by various researchers worldwide, including tracking the hand using glove sensors and artificial vision techniques. Some of the significant works include color segmentation, contour detection and infrared segmentation [2-5]. Recognition algorithms such as searching the fingertips or detecting the valleys between the fingers and fingertips are the ways to extract hand gestures [2, 4]. Over the years, the Microsoft Kinect sensor is a device with a depth sensor. An RGB camera has helped many researchers improve the existing system in terms of high accuracy [6].

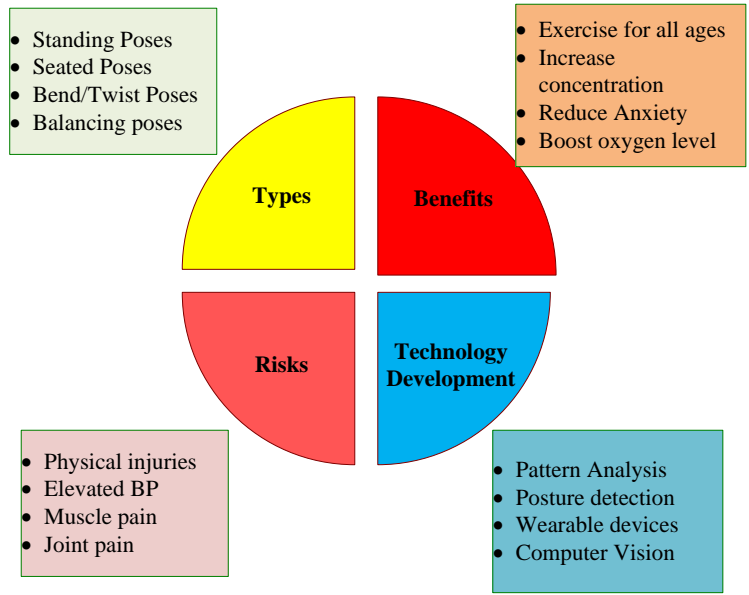

Figure 1: Brief Description of Yoga Activities

Some researchers developed their ideas using image processing with the 4-connected skeleton of the segmentation mask for hand tracking and detection where a camera takes the video in the segmentation part [7]. A mask is extracted according to the user's hand. The noisy points are removed using different morphological operations and the 4-connected skeleton is extracted. In the final part, the center of the hand and position of the fingers are extracted. The work presented by [8] proposed hand tracking using a sum of anisotropic Gaussians model where multiple RGB cameras are used and input images are gathered following some input pre-processing. It combines the 2D Quadtree and 3D SAG models (Automatic model fitting) and then optimizes the pose to arrive at the tracked hand pose. Few researchers have proposed the idea of hand tracking through a single depth camera [9, 10]. An input depth stream can be obtained through any depth camera like Microsoft Kinect or Intel Real Sense, then identifying a square region of interest (RoI) around the hand and segment hand from the background after applying a layered discriminative model to the Region of interest. Optimizing it using a stochastic optimizer provides a flexible setup such as tracking at large distances and over-the-shoulder camera placement.

### 2.1 Motivations from Prior work

According to the Statista survey, in 2018 yoga practitioners has reached 28.75 million. Indeed the numbers will be increasing in future due to the advantages offered by yoga. Due to fulfill the technological gap, such that the traditional trainer-user model will still be in existence and assistance should be available through the mobile phone-based application. The initial implementation of Yoga posture detection presented in [1] was a test based on a dataset developed. It was observed in trials on various users that the mudras i.e. hand gestures are also an essential component in practising yoga. With such motivation, these features have been implemented in the present work.

## 3  Why Yoga is required for Healthcare

Yoga posture recognition explores the use of Microsoft Kinect to identify the posture in real-time. The low-resolution infrared sensor-based work is presented by [4], the use of deep learning algorithms and posture detection in the indoor environment is presented, which shows promising results. Another work presented by [5] is also based on machine learning and deep learning classifiers. The work presented in this work fills the following research gaps:

1. Eliminating the use of external or extra hardware for Yoga posture detection.
2. Execution of Artificial intelligence algorithms on mobile platforms.
3. Real-time assistance with data log features.

# 4    Novel Contribution in the Current Paper

The proposed iYogacare system is explored for real-time yoga recognition and self-correction for health of the human being. This real-time framework is represented as a virtual yoga instructor with minimum error in yoga postures. The research contribution is judiciary highlighted to represent the innovative work. The novel contribution is represented as follows:-

1. Feature points and users daily/weekly activity report are collected using proper standards of protocols to keep track of the postures concerning time. This part of the work is primarily considered for precise posture detection.
2. Information has been collected through audio and text assistance for real-time posture detection.
3. Improvement records and session details as trained module or data have been utilized and shared among the people for precise yoga recognition and self-correction is also possible through these procedures.
4. Mobile application-based platform is used to provide the precise mudras and postures to improve the self yoga activities for individuals. The same application is used to provide the facility of practising the yoga.

# 5    A Novel iYogacare Framework for Healthcare

There is a framework, which plays the role of the instructor. It represents yoga performance poses and also instructs for correction if the person is doing wrong. In prior work, we have performed the detection of the yoga pose with almost 98% accuracy. Now, we have taken this to one level up to where we have added many more features. The first one is the correction feature that helps you to correct the posture if anyone is performing it wrong. This is the main feature as if the instructor is not telling the person that he/she is doing right or wrong then there is no role of instructor at that moment. But, in our present work, there is a feature to provide instructions for correct postures and "Mudras". As we all know that Mudras posture is also equally important as yoga posture. It is a physical activity performed during yoga and it has the potential to produce joy and happiness. It has already been proved that the different alignment of our hands has the power to influence our physical and emotional strength. Mudras help to connect the body to the brain, stimulate endorphins, relieve pain, increase our energy level, and change moods. So this all things kept in the mind. Another feature is added in the framework, which is mudras posture detection. Here, data science and computer vision techniques are used for the detection of postures. A big image data set has been created for detection, which contains 2500 images of 5 different mudras poses. Features have been extracted after creating a hand skeleton. The skeleton has been created by marking all the joints and then connecting them. The angle of the joints has been taken as a feature for different machine learning models. Many machine learning models have been used for detection and among all XGBoost with Ran-domSearch CV models provided the highest accuracy, which is 99% approximately. In Fig.2, it is explained the advantages of doing yoga regularly the benefits include proper blood circulation, relief from joint and muscular pain, healthy heart and lungs etc.

# 6    Methodology for Intelligent Recognition and Self-correction

This feature helps to detect whether a person is doing the posture correct or not and if it follows wrong, then the system will correct and provide different commands. In this work, the tf-pose algorithm is used for extracting different coordinates. By using the coordinates, different angles of joints, distance and slops between two-point have been extracted and used as an evaluation parameter for the correction of the yoga poses. Different Mathematics formulas have been used for the correction such Cosine rule, Euclidean distance. After extracting the parameters, a comparison has been done between the extracted and predefined parameters for each joint and slop for justifying the conclusion of the correctness of the posture. The yoga posture correction is represented in Fig. 3.

# 7    The proposed Framework for iYogacare

## 7.1    Data Collection and Analysis

Mudras are a part of a yogic culture where the practitioners can manipulate energy and are used for therapeutic application. In simple words, mudras mean a gesture created using a different configuration of joining of fingertips. To find a structured dataset with perfect lighting for exclusive yoga mudras is a difficult task. So creating an effective and accurate data set was the first step of the process. This dataset consists of 5 Yoga Mudras namely Pataaka, Mudrakhya, Prana, Pallava and Tripataka Mudra which were captured using a 16 MP, f/1.7, 1/2.8" mobile camera. Each yoga mudra pose consists of 500 images of both left and right hands with a mixed variety of people from different age groups. In total 2500 coloured images are collected in this dataset. Yoga Mudra of five classes is shown in Fig. 5 and 6.

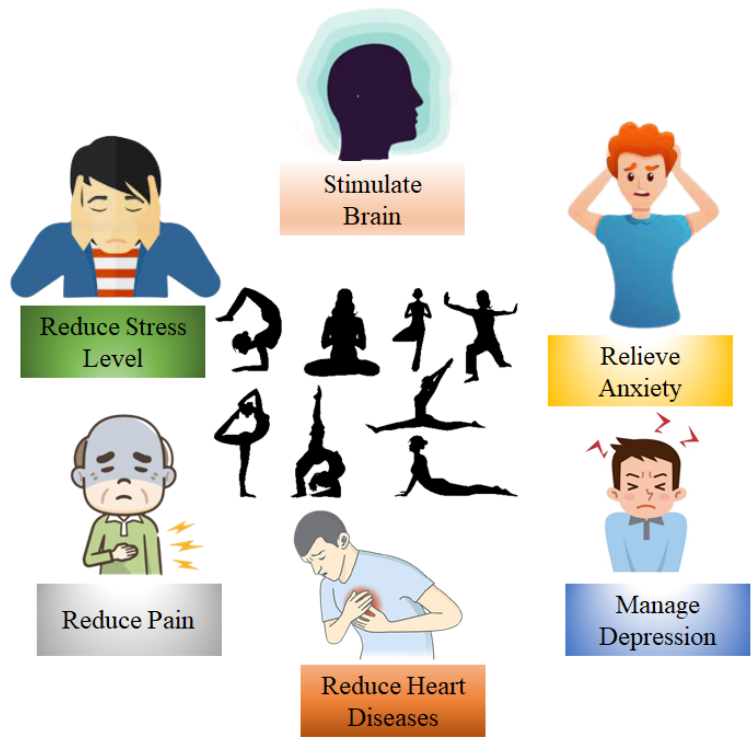

Figure 2: Benefits of Yoga activities on regular basis

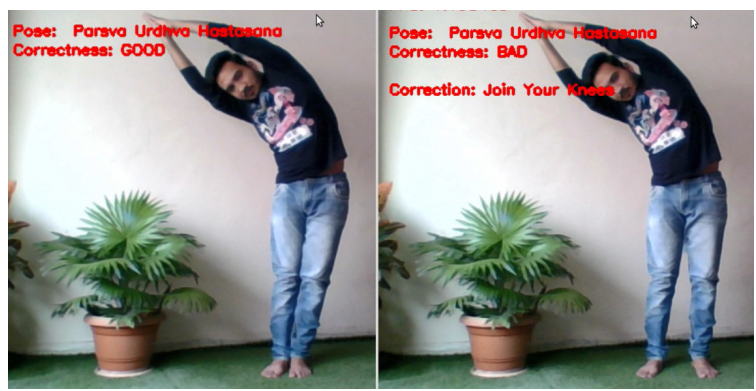

Figure 3: Representation of yoga posture correction

### 7.2 The procedure followed for collecting Mudra dataset

Following steps has been taken after collection mudras dataset. Steps are communicated in Fig 3.

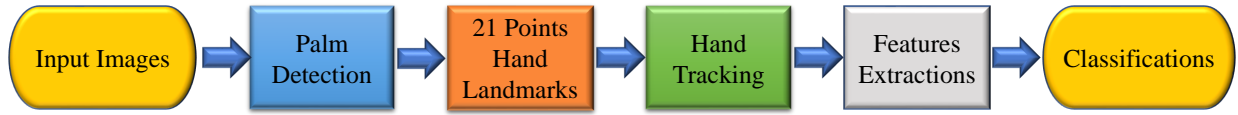

Figure 4: Process flow of posture detection and correction

- Images were taken in different environments such as in a closed room as well as in an open environment so that the final model will work fine in every environment.
- Mobile camera has been used for creating a dataset because, in the end, the software will run as a mobile application.
- The camera has been kept at a distance of around 1m from the hands for perfect images.
- The background was kept white to increase the efficiency of the model while training.
- The images are of both left as well right hand so the software can detect both the hand mudras posture correctly while doing yoga poses.

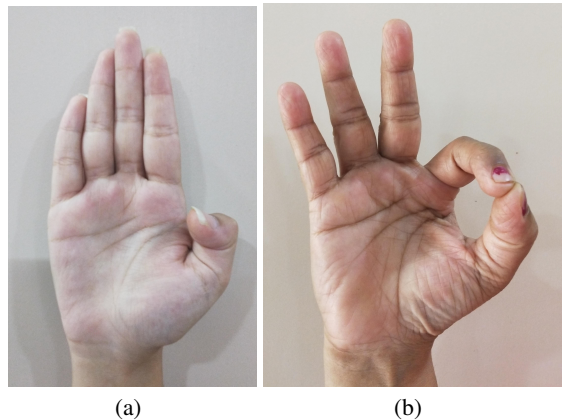

(a)       (b)

Figure 5: Representations of Mudras posture using right hand

## 8 Experimental Analysis and Results

### 8.1 Hand Tracking

Creating a real-time hand tracking, the system is always a difficult computer vision task, as hands often occlude themselves or each other(e.g. handshakes). Google's MediaPipe Hands is a high-fidelity hand and finger tracking solution[11]. This solution locates 21 3D landmarks of the hand using Machine Learning(ML) from a single frame.

A robust ML pipeline has been used with mainly two models working efficiently together. The first model is a palm detector model, which creates a hand bounding box when given an image as a whole. This hand bounding box image is, then treated as an input image to the second model known as a hand landmark model, which returns high fidelity 3D hand keypoints. A Highly efficient solution can be created by accurately providing the and bounding box to the hand landmark model, which helps the system to dedicate its capacity towards predicting the coordinates accurately. It reduces the need for data augmentation (eg: rotation, scale and translation). After that, a hand skeleton is created to track the movement of the hand. A visualization of the hand tracking is shown in Fig 4.

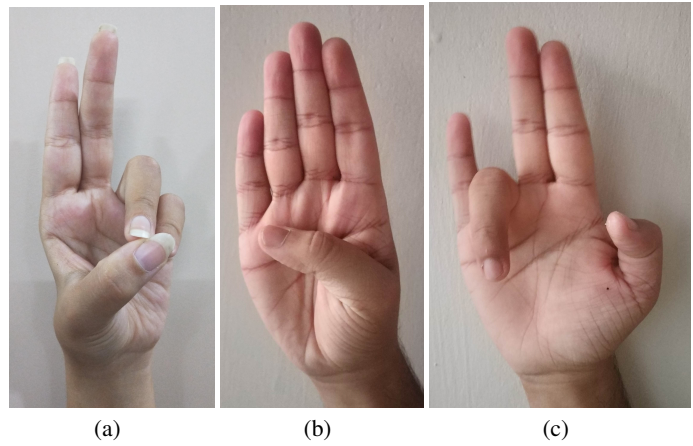

Figure 6: Representations of Mudras posture using left hand

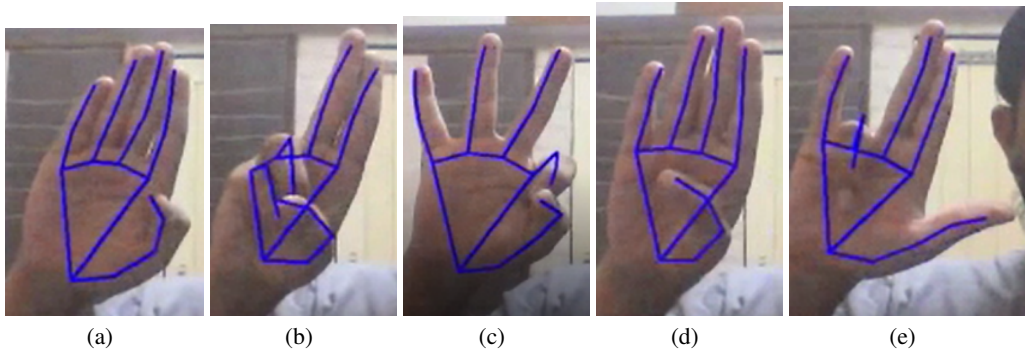

Figure 7: Representations of Mudras posture

### 8.2 Feature Extraction

In the next stage, a skeleton of the palm is created using a media pipe hand tracking system, from which the coordinates of the joints are extracted. The number of joints is shown in Fig. 8. Different angles are extracted using these joint coordinates and then these angles is used as a feature for the detection of hand mudras.

The formula for calculating the distance and angle [13] is shown in Fig. 9.

### 8.3 Classification

The angles extracted from the previous stage, are directly given to the classifier model to predict which yoga mudra is being performed. This happens continuously in real-time. An 80:20 split is performed to bifurcate the data into training and test datasets.

Ten classification models namely k-nearest neighbours (KNN), Random Forest, Shallow NN, Deep NN, OnevsRest, SVM, Logistic Regression, Naive Bayes, XGBoost with RandomSearch CV and Decision Tree were used to benchmark the dataset on various parameters.25 results of accuracy were calculated. XGBoost with RandomSearch CV has the best accuracy and the analysis report is shown in Table-I. The training set consist of 2000 images and the test set consist of 500 images. The following results are mentioned in Table II. A heatmap of the confusion matrix is shown in Fig. 10.

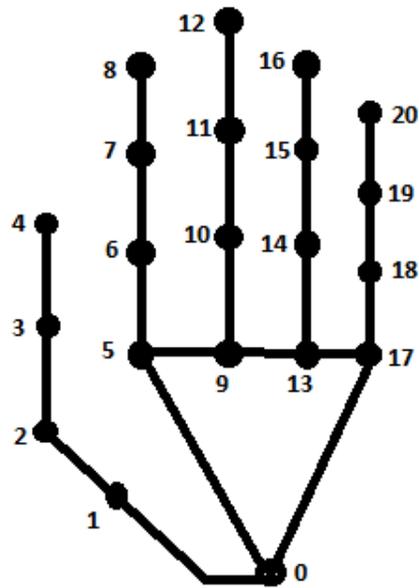

Figure 8: Representation of palm skeleton for features extraction

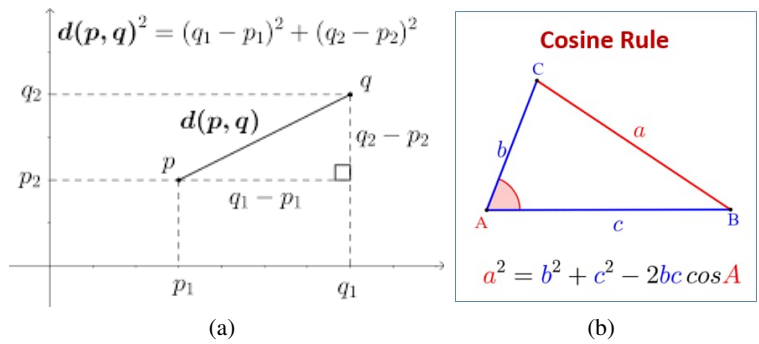

(a)                                       (b)

Figure 9: Representation of formula for calculating the distance and angle

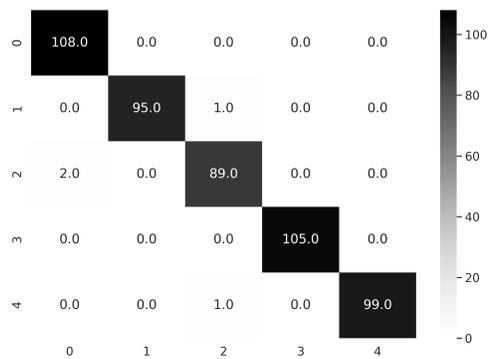

Figure 10: Heatmap of Confusion Matrix

Table 1: Model wise Accuracy for Mudra dataset

| Classifier | Parameters | Accuracy |
|---|---|---|
| k-nearest neighbors (KNN) | neighbors:3, Weight:Uniform, Metric:Minkowski | 0.978 |
| | neighbors:5, Weight:Uniform, Metric:Minkowski | 0.970 |
| | neighbors:9, Weight:Uniform, Metric:Minkowski | 0.964 |
| Random Forest | estimator:30, Criterion:'gini', MaxDepth:7 | 0.984 |
| | estimator:30, Criterion:'gini', MaxDepth:10 | 0.990 |
| | estimator:30, Criterion:'gini', MaxDepth:None | 0.990 |
| Shallow NN | Hidden Layer Size:100, Activation:relu | n0.986 |
| Deep NN | Hidden Layer Size:500, Activation:relu | 0.982 |
| OnevsRest | Hidden Layer Size:500 | 0.990 |
| SVM | Kernal: Linear, Loss Function: Hinge | 0.952 |
| | Kernal: Polynomial, Loss Function:Hinge | 0.978 |
| | Kernal:Radial Basis Function, Loss Function:Hinge | 0.988 |
| Logistic Regression | Iteration Number:2500, Solver:Newton-cg | 0.942 |
| | Iteration Number:2500, Solver:Lbfgs | 0.942 |
| Naive Bayes | Distribution - Normal | 0.920 |
| XGBoost with RandomSearch CV | Booster:GBTree | 0.992 |
| Decision Tree | MinSampleLeaf:1, Splitter:'best', MinSampleSplit:2 | 0.986 |
| | MinSampleLeaf:2, Splitter:'best', MinSampleSplit:2 | 0.972 |
| | MinSampleLeaf:1, Splitter:'best', MinSampleSplit:3 | 0.986 |
| | MinSampleLeaf:2, Splitter:'best', MinSampleSplit:3 | 0.972 |

Table 2: Analysis of XGBoost with RandomSearch CV

| Mudras | Precision | Recall | F1-score | Support |
|---|---|---|---|---|
| Pataaka | 1.00 | 1.00 | 1.00 | 108 |
| Mudrakhya | 0.98 | 0.99 | 0.98 | 96 |
| Prana | 0.98 | 0.98 | 0.98 | 91 |
| Pallava | 1.00 | 1.00 | 1.00 | 105 |
| Tripataka | 1.00 | 0.99 | 0.99 | 100 |

## 9 Conclusion and Future Direction

The system which we have suggested detect and correct 10 different yoga poses and 5 different Mudra poses. The Dataset of mudras poses have been tested on ten algorithms and the highest accuracy we have got is 99.2% on XGBoost with RandomSearch CV. The features from the image have been extracted by finding the coordinate of each joint and then found the angle using that coordinates. The preprocessing of data and training of the model has been done on Ubuntu 18.04.4 LTS terminal and Google Colab. In future, we will extend this project by increasing the dataset of yoga and mudra posture and as well we will be going to adding an audio guidance system. The yoga posture detection can be further implemented along with the Internet of Yoga Things (IoYT) framework. It is required to focus on the identification of the postures in other yoga-like practices such as ballet, climbing, Tai-Chi. In our future work, it is required to consider the multi-users in single frame scenarios such as classrooms and practice halls. To improve the wearable system, there is a need to integrate features like blood pressure monitoring and body temperature through other wearable devices, which can be connected to an existing application program.

## About the Authors

**Abhishek Sharma** is an Assistant Professor in Department of ECE, The LNMIIT, Jaipur, India. Contact him at abhisheksharma@lnmiit.ac.in.

**Yash Shah** has completed his graduation in ECE from The LNMIIT, Jaipur, India. Contact him at 17uec139@lnmiit.ac.in

**Yash Agrawal** has completed his graduation in ECE from The LNMIIT, Jaipur, India, Contact him at 17uec136@lnmiit.ac.in

**Prateek Jain** is an Assistant Professor with the SENSE Department, VIT AP University, Amaravathi (A.P.), India. Contact him at p.jain@ieee.org.